\documentclass[letterpaper, 10 pt, conference]{ieeeconf}  

\IEEEoverridecommandlockouts                              

\usepackage{amsmath} 
\usepackage{amssymb}  
\usepackage{gensymb}
\usepackage{bm}
\usepackage{multirow}

\usepackage{graphicx}
\usepackage{color}
\usepackage{import}
\usepackage{subcaption}
\usepackage{hyperref}
\usepackage{algorithm}
\usepackage[font=small]{caption} 
\usepackage{cite}

\newcommand{\Hcal}{\mathcal{H}}

\DeclareMathOperator*{\argmax}{\arg\!\max}

\newcommand{\assoc}{d}

\newcommand{\meas}{\bm{z}}
\newcommand{\pose}{\bm{x}}
\newcommand{\lm}{\bm{\ell}}

\newcommand{\measset}{\mathbf{Z}}
\newcommand{\assocset}{\mathbf{D}}
\newcommand{\assocvec}{\bm{d}}
\newcommand{\poseset}{\mathbf{X}}
\newcommand{\lmset}{\mathbf{L}}

\newcommand{\SEthree}{\mathrm{SE}(3)}

\newcommand{\reals}{\mathbb{R}}

\newcommand{\measspace}{\mathcal{Z}}

\newcommand{\statespace}{\mathcal{X}}
\newcommand{\lmspace}{\mathcal{L}}

\newcommand{\classspace}{\mathcal{C}}

\title{\LARGE \bf
  Probabilistic Data Association via Mixture Models for \\ Robust Semantic SLAM
}

\author{Kevin Doherty$^1$, David P. Baxter$^{1,*}$, Edward
  Schneeweiss$^{1,2,*}$, and John J. Leonard$^1$
  \thanks{$^{1}$Computer Science and Artificial Intelligence Laboratory (CSAIL),
    Massachusetts Institute of Technology (MIT), Cambridge, MA 02139.
    \texttt{\{kdoherty, baxterdp, jleonard\}@mit.edu}.}
    \thanks{$^{2}$Autonomous Mobile Robotics Laboratory (AMRL), University of
      Massachusetts Amherst (UMass Amherst), Amherst, MA 01003
      \texttt{eschneeweiss@umass.edu}. $^{*}$Equal contributors.}
  \thanks{This work was partially supported by the Office of Naval Research under grant
N00014-18-1-2832. K. Doherty acknowledges support from the NSF Graduate Research
Fellowship Program.}%
}

\begin{document}

\maketitle
\thispagestyle{empty}
\pagestyle{empty}

\begin{abstract}
  Modern robotic systems sense the environment \emph{geometrically}, through
  sensors like cameras, lidar, and sonar, as well as \emph{semantically}, often
  through visual models learned from data, such as object detectors. We aim to
  develop robots that can use all of these sources of information for reliable
  navigation, but each is corrupted by noise. Rather than assume that object
  detection will eventually achieve near perfect performance across the lifetime
  of a robot, in this work we represent and cope with the semantic and geometric
  uncertainty inherent in methods like object detection. Specifically, we model
  data association ambiguity, which is typically non-Gaussian, in a way that is
  amenable to solution within the common nonlinear Gaussian formulation of
  simultaneous localization and mapping (SLAM). We do so by eliminating data
  association variables from the inference process through max-marginalization,
  preserving standard Gaussian posterior assumptions. The result is a
  max-mixture-type model that accounts for multiple data association hypotheses
  as well as incorrect loop closures. We provide experimental results on indoor
  and outdoor semantic navigation tasks with noisy odometry and object detection
  and find that the ability of the proposed approach to represent multiple
  hypotheses, including the ``null'' hypothesis, gives substantial robustness
  advantages in comparison to alternative semantic SLAM approaches.

\end{abstract}

\section{Introduction}


The ability to build and use a map of discrete environmental landmarks to
navigate is one of the greatest strengths of the landmark-based
\emph{simultaneous localization and mapping} (SLAM) paradigm, but hinges
critically on reliable recognition of previously mapped landmarks, i.e. data
association. Consistent data association over long periods of time is vital if
we aim to achieve robust robot navigation in the operational limit as ``time
goes to infinity.'' Unfortunately, long-term data association is substantially
more challenging than short-term data association, since uncertainty in robot
pose may grow large enough that many previously observed landmarks may be
reasonable, albeit erroneous, candidates for a loop closure. For this reason,
any mechanism by which we can associate landmarks \emph{uniquely} is of
interest.

Recent advances in the capabilities and reliability of deep neural networks for
object detection and feature extraction have motivated the use of semantics
jointly to distinguish landmarks, taken to be objects, in the environment, and
infer over time the correct semantic class of each landmark, i.e. semantic SLAM.
However, no detection system can be expected to have perfect accuracy. Rather
than build navigation systems that depend on perfect detection and
classification, we aim to develop methods that can take into consideration the
error characterization of perception systems like neural network-based object
detectors.

\begin{figure}
  \centering
  \includegraphics[width=1.0\columnwidth]{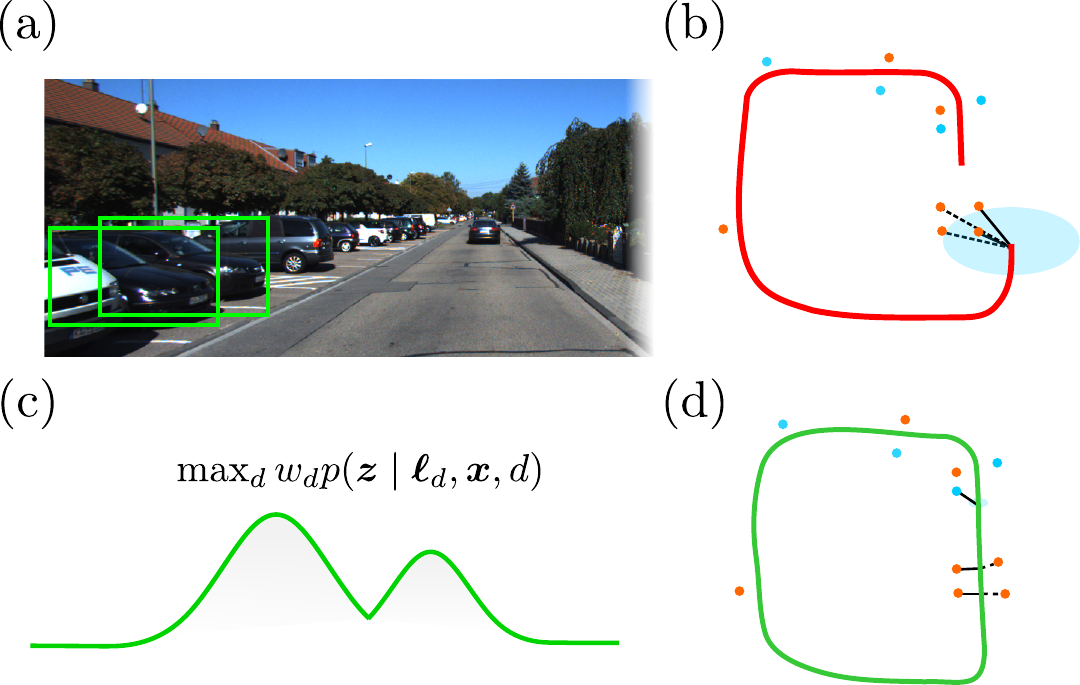}
  \caption{Illustrative overview of the proposed approach. (a) Several object
    detections are made. (b) Robot pose ambiguity (blue covariance ellipse)
    results in multiple candidate landmark associations, the most likely of
    which may be \emph{incorrect}. (c) All candidate associations for a
    measurement are represented as a ``max-mixture'' after data association
    variables are marginalized out. (d) Future evidence allows
    for the association to ``switch'' to the correct hypothesis, fixing the loop
    closure.}
  \label{fig:overview}
\end{figure}

Robustness to misclassification and pose uncertainty requires the abilities to
represent and resolve ambiguity in data association, and to reject incorrect
loop closures. Traditional methods represent the problem of finding the correct
set of \emph{hypotheses} as a tree-search problem, where each node of the tree
represents an association decision. Mitigating the complexity of search requires
careful pruning of plausible hypotheses. Rather than explicitly search over
association hypotheses, in this work we marginalize out data association
variables at each point in time, allowing associations to arise implicitly from
the inference of pose and landmark values.

The main contribution of this work is an \emph{approximate
  max-marginalization} procedure for data associations which provides
theoretical grounding for a ``max-mixture''-type factor
\cite{olson2013inference} within the context of semantic SLAM (shown in Figure
\ref{fig:overview}); thus taking steps toward a unifying perspective on previous
work in ``robust SLAM'' and recent work on data association for semantic SLAM,
e.g. \cite{bowman2017probabilistic}. Approximate max-marginalization eliminates
data association variables in a way that preserves standard Gaussian
distribution assumptions in SLAM in what otherwise becomes a non-Gaussian
inference problem \cite{doherty2019multimodal}. Our representation makes use of
error characterization of an object detector, taking into consideration
uncertainty to fuse detection information with geometric information from other
sensors, like stereo cameras or lidar. Lastly, the proposed method incorporates
loop closure rejection via incorporation of a \emph{null-hypothesis}
association, which we experimentally find to be critical in providing robustness
to odometry noise and misclassification in the semantic SLAM problem.


The remainder of this paper proceeds as follows. In Section
\ref{sect:related-work} we discuss related works on the topics of data
association, robust SLAM, and semantic SLAM. We describe the problem of semantic
SLAM with unknown data association in Section
\ref{sect:semantic-slam-unknown-da}, where we outline our approach to data
association at a high-level. In Section \ref{sect:max-mixture-semantic-slam} we
describe in detail the max-marginalization procedure, data association weight
computation, and define the ``semantic max-mixture factor'' used for
optimization, including the representation of null-hypothesis data association
for loop closure rejection. Finally, experimental results demonstrating the
robustness of the proposed approach to odometry noise and misclassification
during indoor semantic navigation and results from the KITTI dataset
\cite{geiger2012kitti} are provided in Section \ref{sect:experimental-results}.

\section{Related Work}
\label{sect:related-work}

The proposed work intersects the topics of \emph{data association}, \emph{robust
  SLAM}, and \emph{semantic SLAM}. Classical work on data association stems from
target-tracking literature, where probabilistic data association (PDA)
\cite{reid1979algorithm} and multi-hypothesis tracking (MHT)
\cite{bar1975tracking} were introduced. Subsequently these methods were applied
to early filtering-based SLAM solutions making Gaussian noise assumptions
\cite{cox1994modeling, cox1991probabilistic}. FastSLAM
\cite{montemerlo2002fastslam} later introduced a particle filtering-based
approach to the non-Gaussian inference problem of data association; a data
association sampler was introduced that serves as an alternative to explicit
search over associations.

Later work in the area of \emph{pose-graph optimization} focused on themes of
multi-hypothesis SLAM and outlier rejection, including methods like switchable
constraints \cite{sunderhauf2012switchable}, max-mixtures
\cite{olson2013inference}, junction tree inference \cite{segal2014hybrid}, and
robust estimation using convex relaxations \cite{carlone2014selecting,
  carlone2018convex, lajoie2019modeling}. These works consider mitigating the
effects of perceptual aliasing, often in the context of laser scan matching or
appearance-based loop closure (see, for example \cite{lowry2015visual}), whereas
in semantic SLAM we also want to locate and classify discrete objects.
Nonetheless, as we demonstrate in this work, landmark-based semantic SLAM shares
similar challenges. Specifically, we take an approximate Bayesian inference
perspective on the semantic SLAM problem and arrive at a specific case of the
max-mixtures method \cite{olson2013inference} where component weights are
directly computed as candidate association probabilities.

In the area of semantic SLAM, classifications from object detectors have been
used to aid data association \cite{bowman2017probabilistic, doherty2019multimodal, yang2018cubeslam,
  salas2013slam++, mu2016slam}. While many such works
consider maximum-likelihood data association \cite{yang2018cubeslam,
  salas2013slam++}, recent works have considered probabilistic data association
making use of expectation-maximization \cite{bowman2017probabilistic}, or
alternate between sampling data associations and recomputing SLAM solutions
\cite{mu2016slam}. In contrast to \cite{bowman2017probabilistic}, we model data
associations as a mixture, rather than averaging solutions for different
associations. We also marginalize out poses and landmarks when computing data
association probabilities, whereas in \cite{bowman2017probabilistic}, point
estimates of robot poses and landmarks are used to compute data association
weights. In both \cite{bowman2017probabilistic} and \cite{mu2016slam},
convergence requires iteratively recomputing data associations and performing
factor graph optimization; we aim to avoid the complexity associated with this
recomputation. In previous work \cite{doherty2019multimodal}, we addressed this
problem for the general non-Gaussian SLAM case using nonparametric belief
propagation. While that approach gives rich uncertainties representing data
association ambiguity, in this work our goal is maximum \emph{a posteriori}
inference specifically in the nonlinear Gaussian case.

\section{Semantic SLAM with Unknown Data Association}
\label{sect:semantic-slam-unknown-da}

We define the semantic SLAM problem to be the inference of vehicle
poses $\poseset = \{\pose_t \in \statespace \}_{t=1}^T$ and landmark states
$\lmset = \{\lm_j \in \lmspace\}_{j=1}^M$, given a set of measurements
$\{\meas_t \in \measspace\}_{t=1}^T$. This corresponds to the following maximum
\emph{a posteriori} (MAP) inference problem:
\begin{align}
  \hat{\poseset}, \hat{\lmset} &= \argmax_{\poseset, \lmset} p(\poseset, \lmset \mid \measset).
\end{align}
We consider the vehicle state space $\statespace \triangleq \SEthree$
throughout, while we consider landmark states containing both geometric and
semantic components, i.e. $\lmspace \triangleq \reals^3 \times \classspace$ (for
landmarks with only a positional component) or $\lmspace \triangleq \SEthree
\times \classspace$ (for 6 degree-of-freedom landmark pose estimation), where
$\classspace = \{1, \ldots, C\}$ is a fixed, \emph{a priori} known set of $C$
discrete semantic classes. When necessary, we denote the separate
pose/positional components and semantic components of a landmark $\lm_j$ as
$\lm_j^p$ and $\lm_j^s$, respectively. We assume multiple measurements can be
made at each point in time, such that $\meas_{tk}$ is the $k$-th measurement made
at time $t$. Lastly, we consider the measurement space $\measspace$ as having
jointly geometric and semantic components, e.g. range-bearing measurements with
semantic class in $\classspace$ or 6-DoF pose measurements in $\SEthree$ with
semantic class in $\classspace$.

When associations between measurements and landmarks are not known, they must be
inferred. Specifically we take $\assocset = \{\assocvec_t\}_{t=1}^T$ to be the
set of associations of measurements $\meas_t$ at all points in time to
landmarks, such that $\assoc_{tk} = j$ indicates that the $k$-th measurement
taken at time $t$ corresponds to landmark $\lm_j$. The most common approach to
SLAM with unknown data association is that of \textit{maximum-likelihood}, in
which the most probable set of data associations are computed and fixed, then
used to solve for the most probable robot poses and landmark states. This
approach can be brittle, as a single incorrect association can move the optimal
solution for robot poses and landmark states far from their true values.

In order to mitigate the effects of data association errors one may consider
\textit{probabilistic data association}, in which multiple associations for a
measurement are given consideration commensurate with their probability.
Generally this corresponds with marginalization of the data association
variable, i.e.
\begin{align}
  p(\poseset, \lmset \mid \measset) &= \mathbb{E}_{\assocset}\left[ p(\poseset, \lmset \mid \assocset, \measset) \mid \measset \right] \label{eq:slam-da-marg} \\
                                    &= \sum_{\assocset} p(\poseset, \lmset \mid \assocset, \measset)p(\assocset \mid \measset). \nonumber
\end{align}
Common assumptions of additive Gaussian measurement noise make $p(\poseset,
\lmset \mid \assocset, \measset)$ Gaussian, such that the resulting belief
$p(\poseset, \lmset \mid \measset)$ is a sum of Gaussians, which generally falls
outside the realm of traditional nonlinear least-squares optimization approaches
to SLAM.

A recent approach to semantic SLAM \cite{bowman2017probabilistic} preserves the
Gaussian nature of the problem by replacing the above expectation with one over
$\log p(\poseset, \lmset \mid \assocset, \measset)$, leading to an
expectation-maximization algorithm for optimization. Convergence requires
recomputation of the association weights $p(\assocset \mid \measset)$, and prior
to convergence the solution will lie somewhere \emph{between} those obtained
given fixed associations.

We propose an alternative solution to the MAP inference problem in which the
``sum-marginal'' above is replaced by the ``max-marginal'':
\begin{align}
  \hat{p}(\poseset, \lmset \mid \measset) \triangleq \max_{\assocset} p(\poseset, \lmset \mid \assocset, \measset)p(\assocset \mid \measset). \label{eq:max-marginal}
\end{align}
Each component of the max-marginal $\hat{p}(\poseset, \lmset \mid \measset)$ is
a weighted Gaussian, while the $\max$ operator simply acts to switch to the
``best'' data associations for any given point in the latent space of $\poseset$
and $\lmset$. The optimal solution for the max-marginal is identical to the MAP
solution for the true posterior in \eqref{eq:slam-da-marg}
\cite{koller2009probabilistic}. Exact computation of the true max-marginal is
generally intractable due to the combinatorial number of plausible data
associations, but as we will show, several reasonable approximations make the
max-marginal a practical method for dealing with data association ambiguity.

\begin{figure}
  \centering
  \begin{subfigure}{\columnwidth}
    \centering
    \def\svgwidth{\columnwidth}
    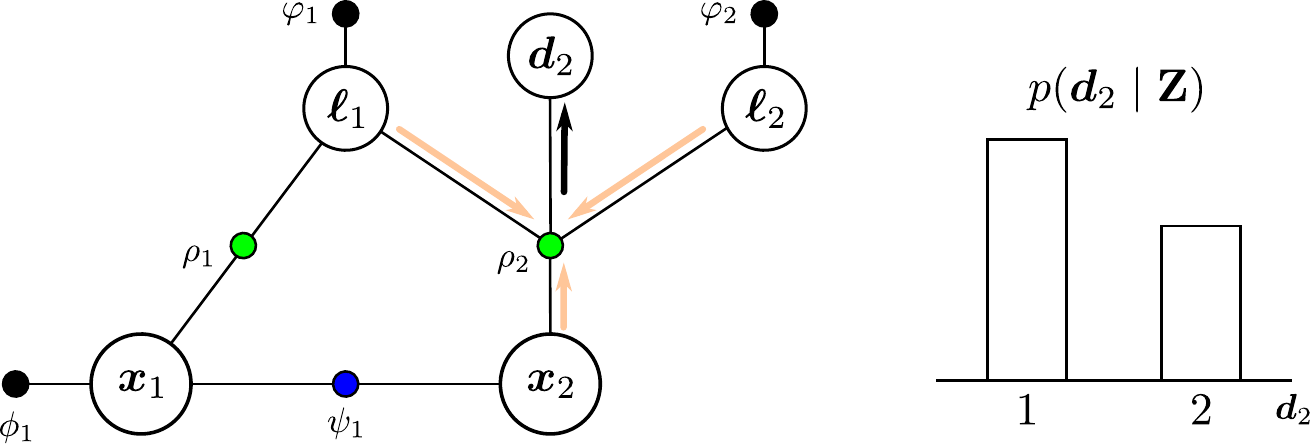
    \caption{Association probabilities computed by marginalizing out poses and landmarks.\label{fig:da-example:probs}}
  \end{subfigure}

  \begin{subfigure}{\columnwidth}
    \centering
    \def\svgwidth{\columnwidth}
    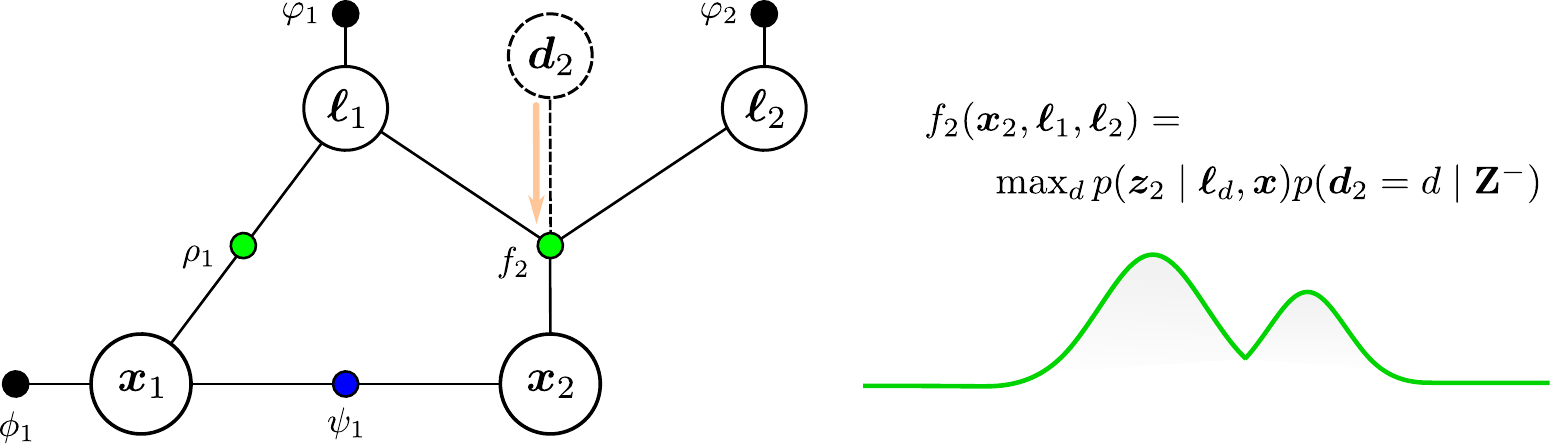
    \caption{Data association max-marginalization produces a mixture factor.\label{fig:da-example:mixture}}
  \end{subfigure}
  \caption{Illustrative example of data association variable marginalization. In
    \ref{fig:da-example:probs}, we marginalize out poses and landmarks to
    compute a distribution over candidate data associations (see Section
    \ref{sect:da-weights}). In \ref{fig:da-example:mixture}, we compute the
    max-marginal over data associations, resulting in a max-mixture factor (see
    Section \ref{sect:mixture-factor}).\label{fig:da-marg-process}}
\end{figure}

\section{Max-Mixture Semantic SLAM}
\label{sect:max-mixture-semantic-slam}
We consider a standard Gaussian SLAM framework with an odometry model $p(\pose_t
\mid \pose_{t-1})$ that is Gaussian with covariance $\Sigma_t$ with respect to
the relative transform from $\pose_{t-1}$ to $\pose_t$, denoted $g(\pose_{t},
\pose_{t-1})$ and Gaussian geometric measurement model $p(\meas_{tk}^p \mid
\pose_t, \lm_j)$ with covariance $\Gamma$, with respect to the nonlinear
function $h(\pose_t, \lm_j)$\footnote{The function $h$ in this work is taken to
  be a relative transform in the case of full landmark pose measurements, or
  bearing, elevation, and range when only landmark position information is
  available.}. Semantic measurements with model $p(\meas_{tk}^s \mid \lm_j^s)$
are assumed independent of the pose from which a landmark was observed, as well
as its position, and as in \cite{doherty2019multimodal, bowman2017probabilistic}
they are taken as samples from a categorical distribution with probability
vector defined by a classifier confusion matrix (assumed to be known \emph{a
  priori}). Lastly, we assume the geometric and semantic measurements factor as
$p(\meas_{tk} \mid \pose_t, \lm_j^p) = p(\meas_{tk}^p \mid \pose_t,
\lm_j)p(\meas_{tk}^s \mid \lm_j^s)$.

The unnormalized posterior $p(\poseset, \lmset \mid \measset)$, can be written
in the following general \emph{factor graph} formulation:
\begin{align}
  p(\poseset, \lmset \mid \measset) \propto \prod_i f_i(\mathbf{V}_i),\ \mathbf{V}_i \subseteq \{\poseset, \lmset\},
\end{align}
where each factor $f_i$ is in correspondence with one of the relevant (odometric
or landmark) measurement models. From the measurement models, there is a clear
partition of \emph{geometric} information from the odometry and geometric
landmark measurement models (which depend on both the robot poses and landmark
locations) and the \emph{semantic} information, which depends only on the class
of the associated landmark. Since we do not know data associations, we instead
infer them from data. At a high level, our approach is to apply variable
elimination to data associations to produce an equivalent factor graph with data
associations marginalized out. The proposed \emph{max-mixture semantic SLAM}
approach approximates optimization over the max-marginal in
\eqref{eq:max-marginal}. In particular, we introduce a \textbf{proactive
  max-marginalization} procedure for \textbf{computing data association
  weights}. For associations to previous landmarks, as well as for the
null-hypothesis case, the max-marginal over candidate associations is
represented as a \emph{factor} (in the factor graph SLAM framework) taking on
the form of a ``max-mixture'' \cite{olson2013inference}. We term these factors,
\textbf{semantic max-mixture factors}. The addition of
\textbf{null-hypothesis data association} enables the rejection of
incorrect loop closures. The resulting factor graph is amenable to optimization
using standard nonlinear least-squares techniques, from which the optimal robot
and landmark states can be recovered.

\subsection{Proactive Max-Marginalization}
Exact computation of the max-marginal over all possible data associations in
Equation \eqref{eq:max-marginal} is computationally expensive due to the
combinatorial growth in the size of the set of possible data associations over
time. Marginalizing out data associations \emph{proactively}, i.e. as new measurements
are made, and ignoring the influence of future measurements on association
probabilities allows us to mitigate the complexity of full max-marginalization.

In particular, suppose we have some set of previous measurements $\measset^{-}$
and new measurements $\measset^{+}$ . We aim to compute the max-marginal over
associations to the new measurements, denoted $\assocset^{+}$. Formally, we have
the following:
\begin{multline}
  p(\poseset, \lmset, \assocset^{+} \mid \measset^{+}, \measset^{-}) \propto \\ p(\measset^{+} \mid \poseset, \lmset, \assocset^{+}) p(\poseset, \lmset \mid \measset^{-})p(\assocset^{+} \mid \measset^{-}),
\end{multline}
from Bayes' rule, where we have used the conditional independences
$p(\measset^{+} \mid \poseset, \lmset, \assocset^{+}, \measset^{-}) =
p(\measset^{+} \mid \poseset, \lmset, \assocset^{+})$, and $p(\poseset, \lmset
\mid \assocset^{+}, \measset^{-}) = p(\poseset, \lmset \mid \measset^{-})$,
since $\assocset^{+}$ consists of associations to only measurements outside of
$\measset^{-}$. Applying max-marginalization to data associations, we obtain:
\begin{multline}
  \hat{p}(\poseset, \lmset \mid \measset^{+}, \measset^{-}) = \\
  p(\poseset, \lmset \mid \measset^{-}) \max_{\assocset^{+}} \left[ p(\measset^{+} \mid \poseset, \lmset, \assocset^{+}) p(\assocset^{+} \mid \measset^{-}) \right].
\end{multline}
Here $p(\poseset, \lmset \mid \measset^{-})$ is the (potentially non-Gaussian)
posterior distribution over poses and landmarks after sum-marginalization of
data associations to the measurements $\measset^{-}$. For the purposes of
optimization in the Gaussian case, we take this as the max-marginal
$\hat{p}(\poseset, \lmset \mid \measset^{-})$.

The consequences of this simple change are significant: evaluating the
$\max$ operators in the above expression no longer requires examination of
previous associations and can be done in linear time for the most recent
measurement. The result is that we have arrived at an approximate max-product
algorithm for SLAM with unknown data associations.

\begin{figure}
  \centering
  \begin{subfigure}{1.0\linewidth}
    \centering
    \includegraphics[width=0.5\linewidth]{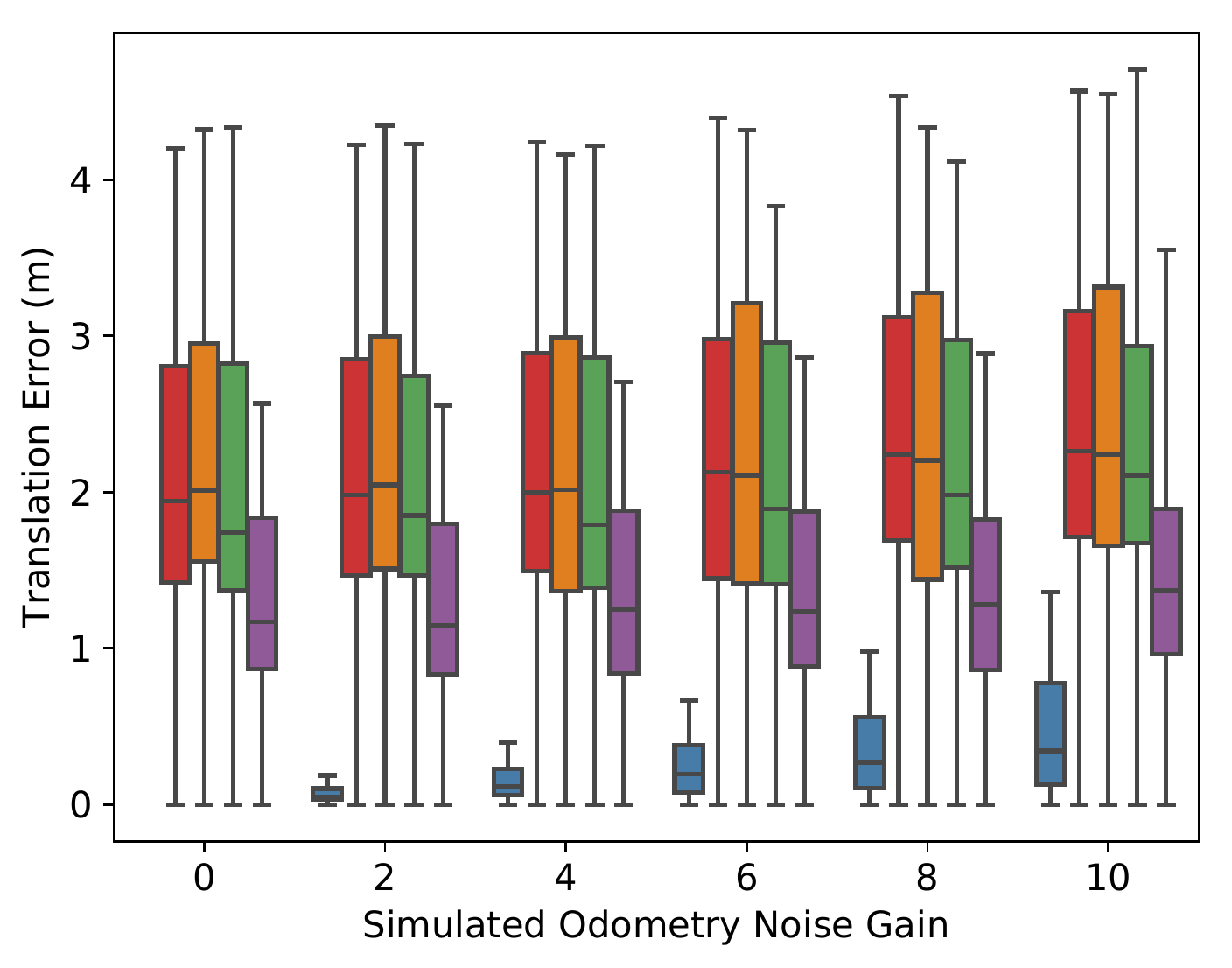}%
    \includegraphics[width=0.5\linewidth]{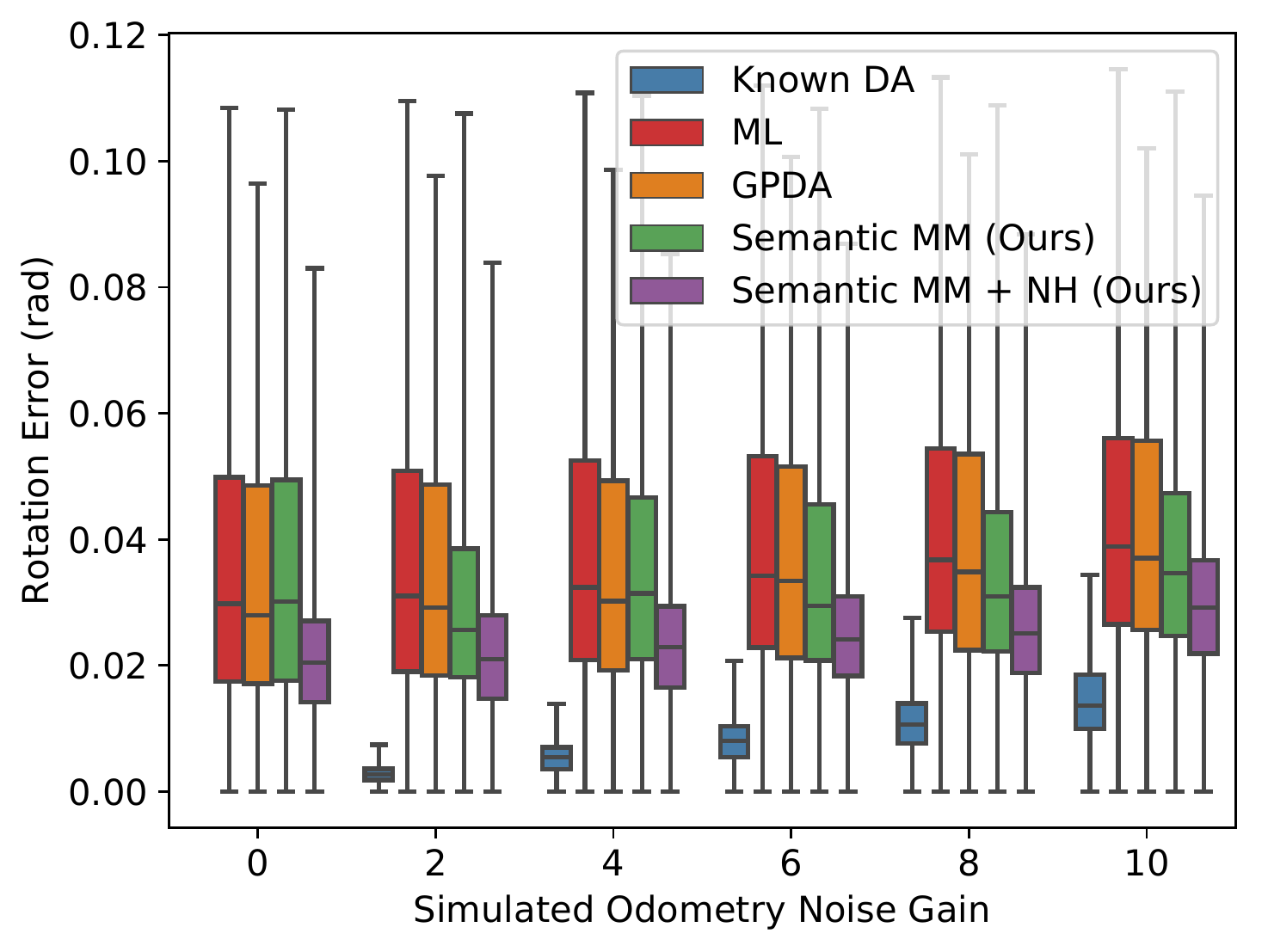}
    \caption{} 
  \end{subfigure}%

  \begin{subfigure}{1.0\linewidth}
    \includegraphics[width=0.5\linewidth]{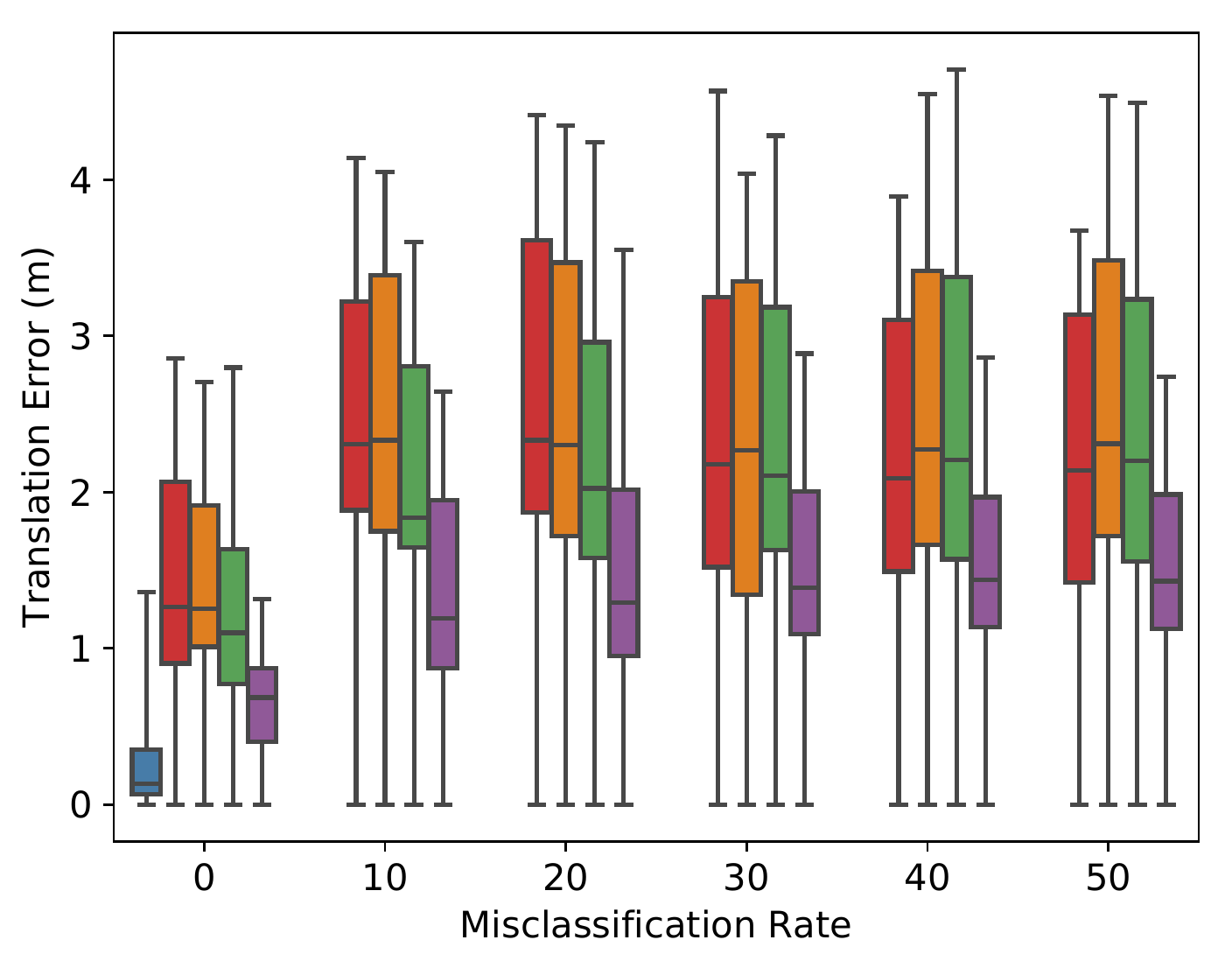}%
    \includegraphics[width=0.5\linewidth]{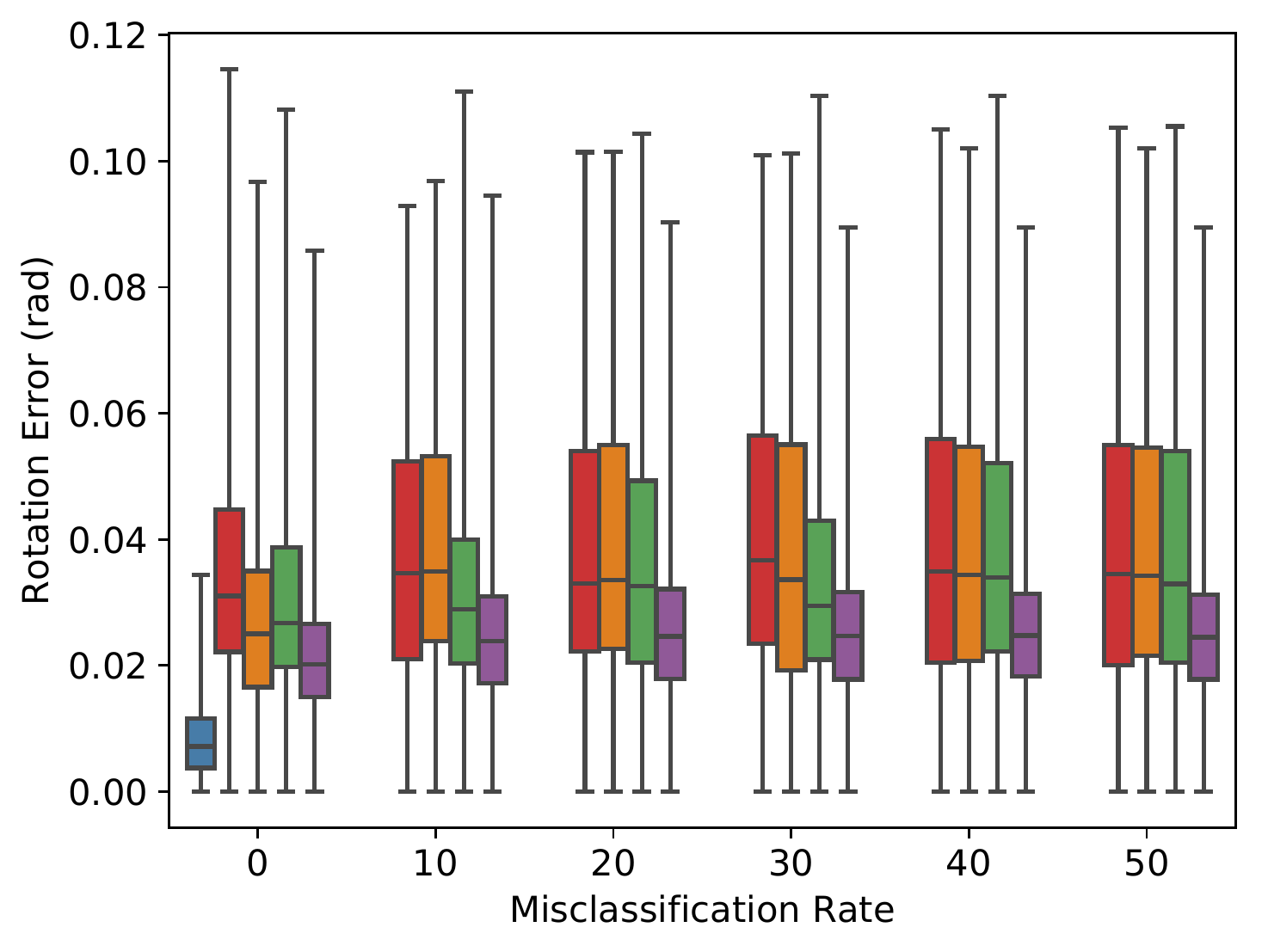}
    \caption{}
  \end{subfigure}%
  \caption{Translation and rotation error as a function of (a) the simulated
    odometry noise and (b) misclassification rate parameters. The probabilistic
    approaches generally outperform maximum-likelihood, but the most significant
    advantages come from the incorporation of the null-hypothesis into the
    mixture-based framework, which outperforms all other methods across all
    misclassification rates and odometry noise. Legend is consistent
    throughout. \label{fig:apriltag-error}}
\end{figure}

\subsection{Data Association Weight Computation}
\label{sect:da-weights}

Consider a single measurement of a landmark $\meas$, which in our case consists
of the joint geometric and semantic measurement of the landmark. We assume the
data association probability is proportional to the likelihood $p(\meas_{tk}
\mid \assoc_{tk}, \measset^{-})$ with poses and landmarks marginalized out (see
Figure \ref{fig:da-example:probs}). From the factored measurement model
assumption, the likelihood of the form $p(\meas_{tk} \mid \assoc_{tk}, \measset^{-})$
can be broken into the product of separate semantic and geometric likelihoods:
\begin{multline}
  p(\meas_{tk} \mid \assoc_{tk} = j, \measset^{-}) = \\ p(\meas_{tk}^s \mid \assoc_{tk} = j, \measset^{-})p(\meas_{tk}^p \mid \assoc_{tk} = j, \measset^{-}). \label{eq:factored-da-likelihood}
\end{multline}
Each term on the right-hand side can be expanded as follows into the
summation over landmark classes:
\begin{multline}
  p(\meas_{tk}^s \mid \assoc_{tk} = j, \measset^{-}) =\\ \sum_{c} p(\meas_{tk}^s \mid \lm_j^s = c)p(\lm_j^s = c \mid \measset^{-}), \label{eq:semantic-da-likelihood}
\end{multline}
and integral over robot pose and landmark location:
\begin{multline}
  p(\meas_{tk}^p \mid \assoc_{tk} = j, \measset^{-}) = \\ \iint p(\meas_{tk}^p
  \mid \assoc_{tk} = j, \pose_t, \lm_j) p(\pose_t, \lm_j \mid
  \measset^{-})d\pose_td\lm_j. \label{eq:geometric-da-likelihood}
\end{multline}
With data associations marginalized out, the belief $p(\pose_t, \lm_j \mid
\measset^{-})$ would be generally non-Gaussian. Consequently, we again make an
approximation and use the single Gaussian component corresponding to the max-marginal
$\hat{p}(\pose_t, \lm_j \mid \measset^{-})$ evaluated at the current estimate of
$\pose_t$ and $\lm_j$, ) which we denote $\hat{p}(\hat{\pose}_t, \hat{\lm}_j
\mid \measset^{-})$:
\begin{multline}
  p(\meas_{tk}^p \mid \assoc_{tk} = j, \measset^{-}) \approx \\ \iint p(\meas_{tk}^p \mid \assoc_{tk} = j, \pose_t, \lm_j) \hat{p}(\hat{\pose}_t, \hat{\lm}_j \mid \measset^{-})d\pose_td\lm_j.
\end{multline}
Since all of the terms in the integral are now Gaussian, it can be simplified as
follows, based on the method of \cite{kaess2009covariance}:
\begin{multline}
  p(\meas_{tk}^p \mid \assoc_{tk} = j, \measset^{-}) \approx \\
  \frac{1}{\sqrt{| 2 \pi R_{tjk}|}} e^{-\frac{1}{2} \| h(\hat{\mathbf{x}}) - \meas_{tk}^p \|_{R_{tjk}}^2},
\end{multline}
where $\hat{\mathbf{x}}$ is the mean of the joint distribution over $\pose_t$ and
$\lm_j$. The covariance $R_{tjk}$, is defined as:
\begin{align}
  R_{tjk} &\triangleq \frac{\partial h}{\partial \mathbf{x}}\bigg\rvert_{\hat{\mathbf{x}}} \Sigma \frac{\partial h}{\partial \mathbf{x}}\bigg\rvert_{\hat{\mathbf{x}}}^T + \Gamma,
\end{align}
where $\Sigma$ is the block joint covariance matrix between pose $\pose_t$ and
candidate landmark $\lm_j$, $\partial h / \partial \mathbf{x}$ is the Jacobian
of the measurement function, and $\Gamma$ is the covariance of the geometric
measurement model. This result, combined with the expression in
\eqref{eq:semantic-da-likelihood} gives the marginal likelihood in
\eqref{eq:factored-da-likelihood} that we normalize to compute data association
probabilities.








\begin{figure}[t]
  \centering
  \begin{subfigure}{0.5\linewidth}
    \centering
    \includegraphics[width=1.0\linewidth]{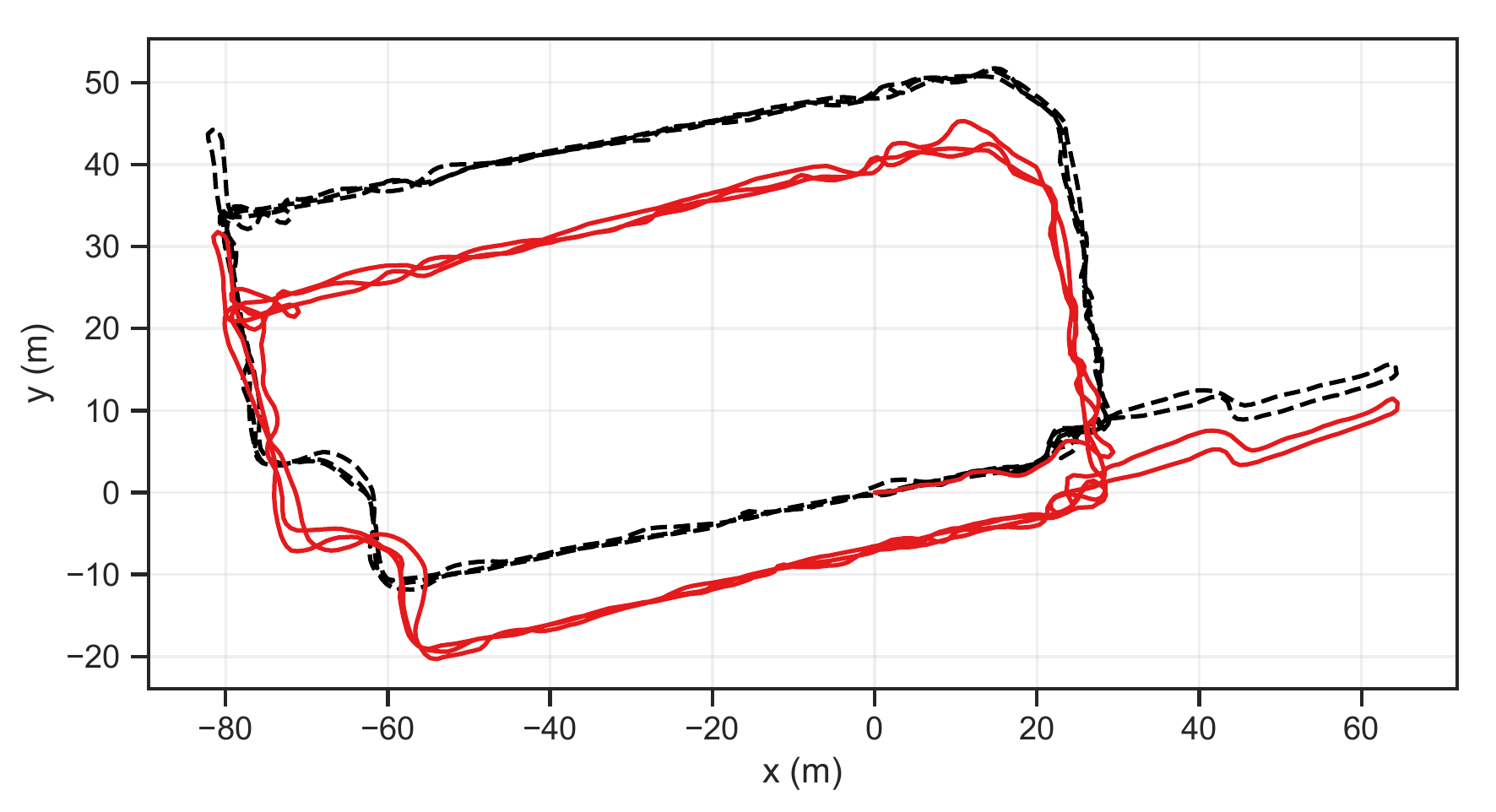}
    \caption{Max-Likelihood \label{fig:apriltag-traj_error-ml}}
  \end{subfigure}%
  \begin{subfigure}{0.5\linewidth}
    \centering
    \includegraphics[width=1.0\linewidth]{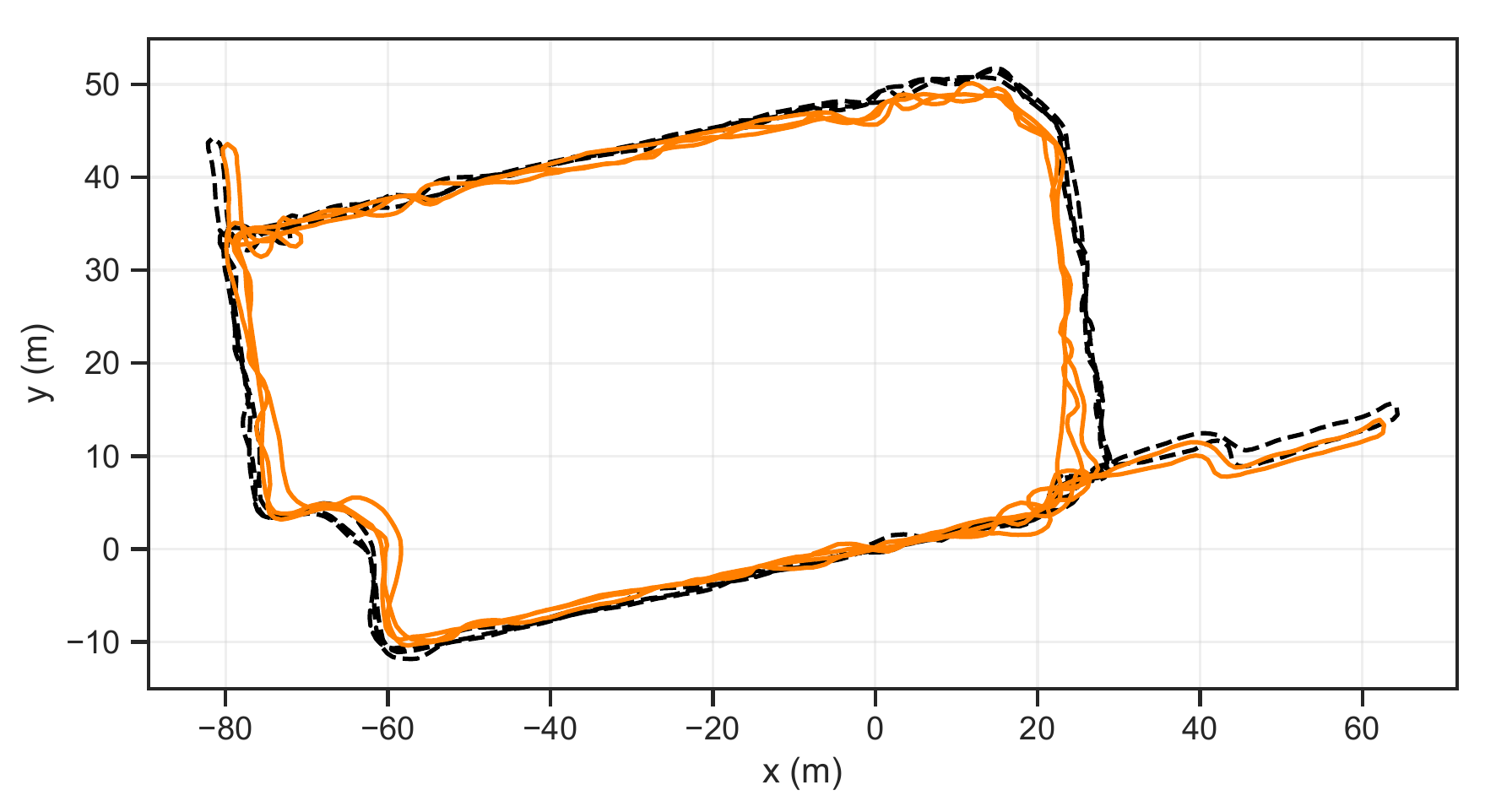}
    \caption{Gaussian PDA \label{fig:apriltag-traj_error-gpda}}
  \end{subfigure}
  \begin{subfigure}{0.5\linewidth}
    \centering
    \includegraphics[width=1.0\linewidth]{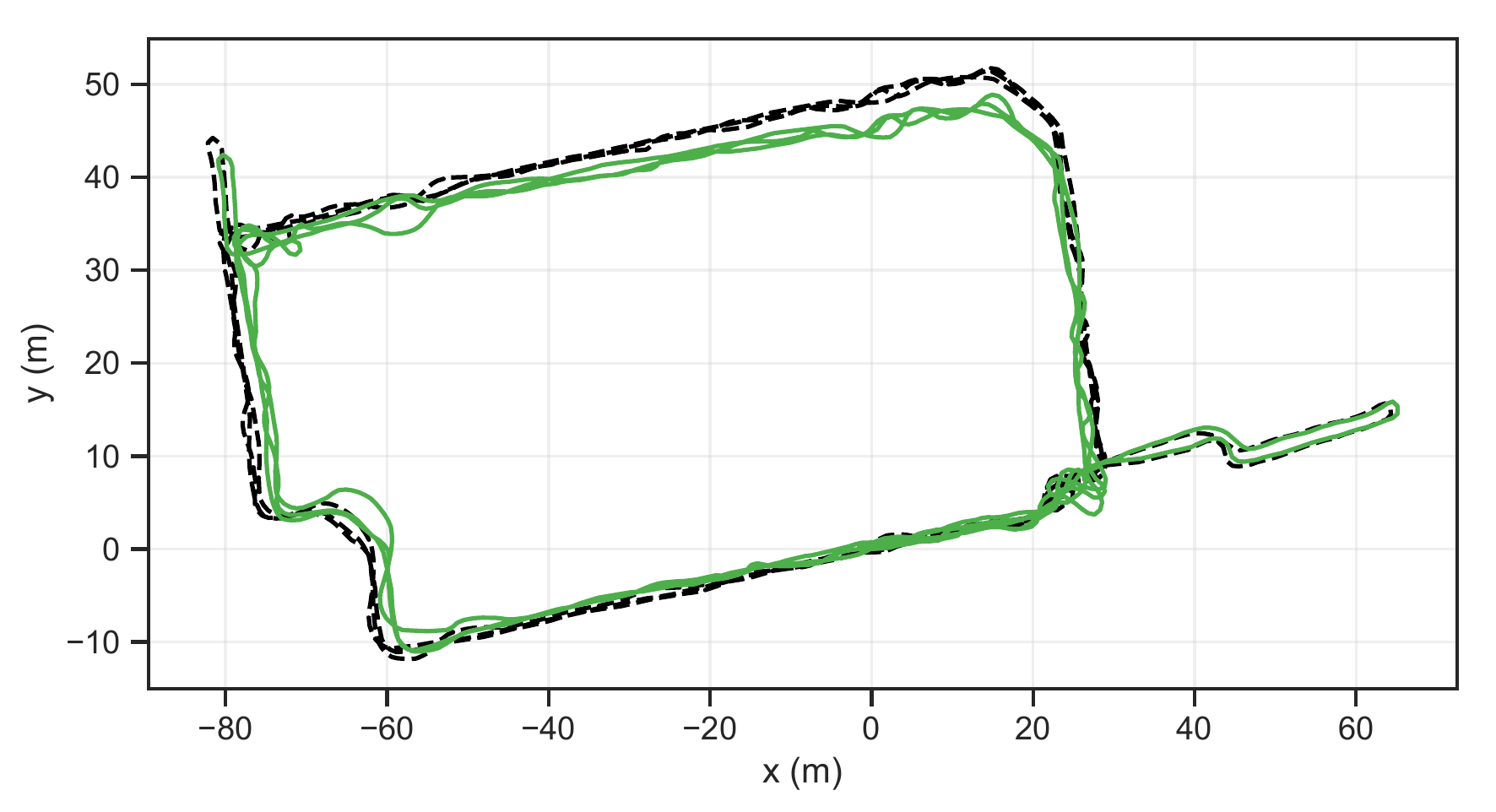}
    \caption{Semantic MM (Ours) \label{fig:apriltag-traj_error-maxmixtures}}
  \end{subfigure}%
  \begin{subfigure}{0.5\linewidth}
    \centering
    \includegraphics[width=1.0\linewidth]{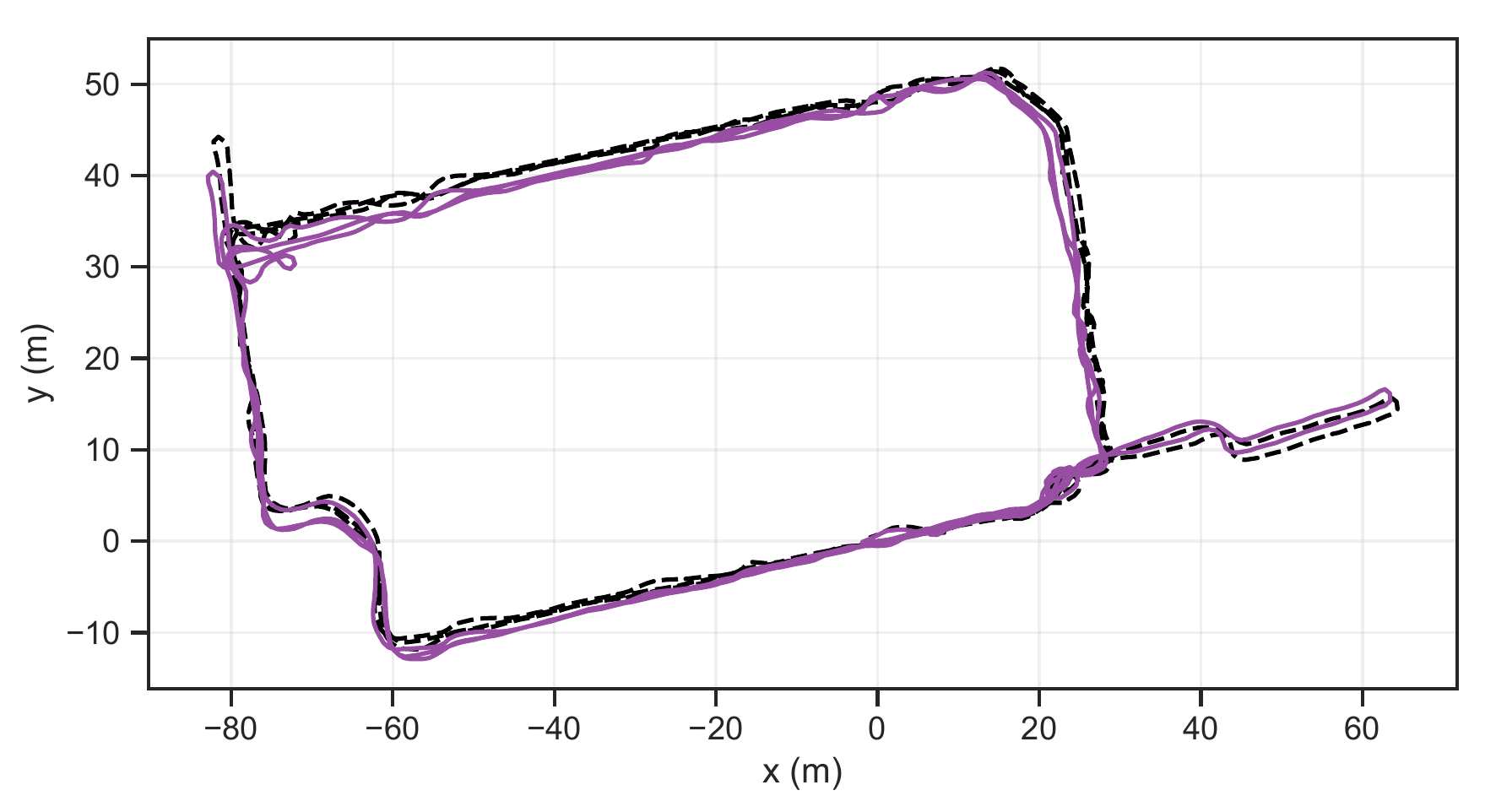}
    \caption{Semantic MM + NH (Ours) \label{fig:apriltag-traj_error-maxmixtures}}
  \end{subfigure}
  \caption{Estimated trajectories for the MIT RACECAR dataset (origin aligned);
    reference trajectory in black. Association errors cause the
    maximum-likelihood approach to produce additional
    hallways. \label{fig:apriltag-traj_error}}
\end{figure}

\subsection{Semantic Max-Mixture Factor}
\label{sect:mixture-factor}
Assuming uniform priors on data associations, the distribution $p(\assoc_{tk} \mid
\measset^{-})$ is proportional to the marginal likelihood in
\eqref{eq:factored-da-likelihood} and can simply be normalized over all
assignments to $\assoc_{tk}$. This results in a set $\Hcal \subseteq \{1, \ldots,
M\}$ of candidate landmark hypotheses, max-marginalization of which produces a
max-mixture factor for a measurement $\meas_{tk}$:
\begin{align}
  f(\pose_t, \lm_{\Hcal}) &= \max_{j \in \Hcal} p(\meas_{tk} \mid \pose_t, \lm_j)p(\assoc_{tk} = j \mid \measset^{-}).
\end{align}
The max-marginalization step is visualized in Figure
\ref{fig:da-example:mixture}, where we have eliminated the data association
variable from the inference process. By augmenting the candidate set $\Hcal$ to
be $\Hcal \cup \{\emptyset\}$, we allow a \emph{null-hypothesis} data
association to be made. In practice, we assume a probability for the
null-hypothesis and normalize the remaining data associations such that the
total probability of the augmented hypothesis set equals 1. The null-hypothesis
component is assumed to be Gaussian with large standard deviation (e.g. $10^5$).

Finally, we can recover maximum \emph{a posteriori} landmark semantic class estimates
(assuming uniform priors) as in \cite{bowman2017probabilistic} as follows:
\begin{align}
  \hat{\lm}_j^s = \argmax_c \prod_t\sum_{\assocvec_t} p(\assocvec_t, \lm_j^s = c \mid \measset), \label{eq:lm-class-map}
\end{align}
which are recovered using the data association probabilities stored as the
component weights of the max-mixture factors.

\section{Experimental Results}
\label{sect:experimental-results}

\begin{figure*}
  \centering
  \begin{subfigure}{0.24\linewidth}
    \centering
    \includegraphics[width=1.0\linewidth]{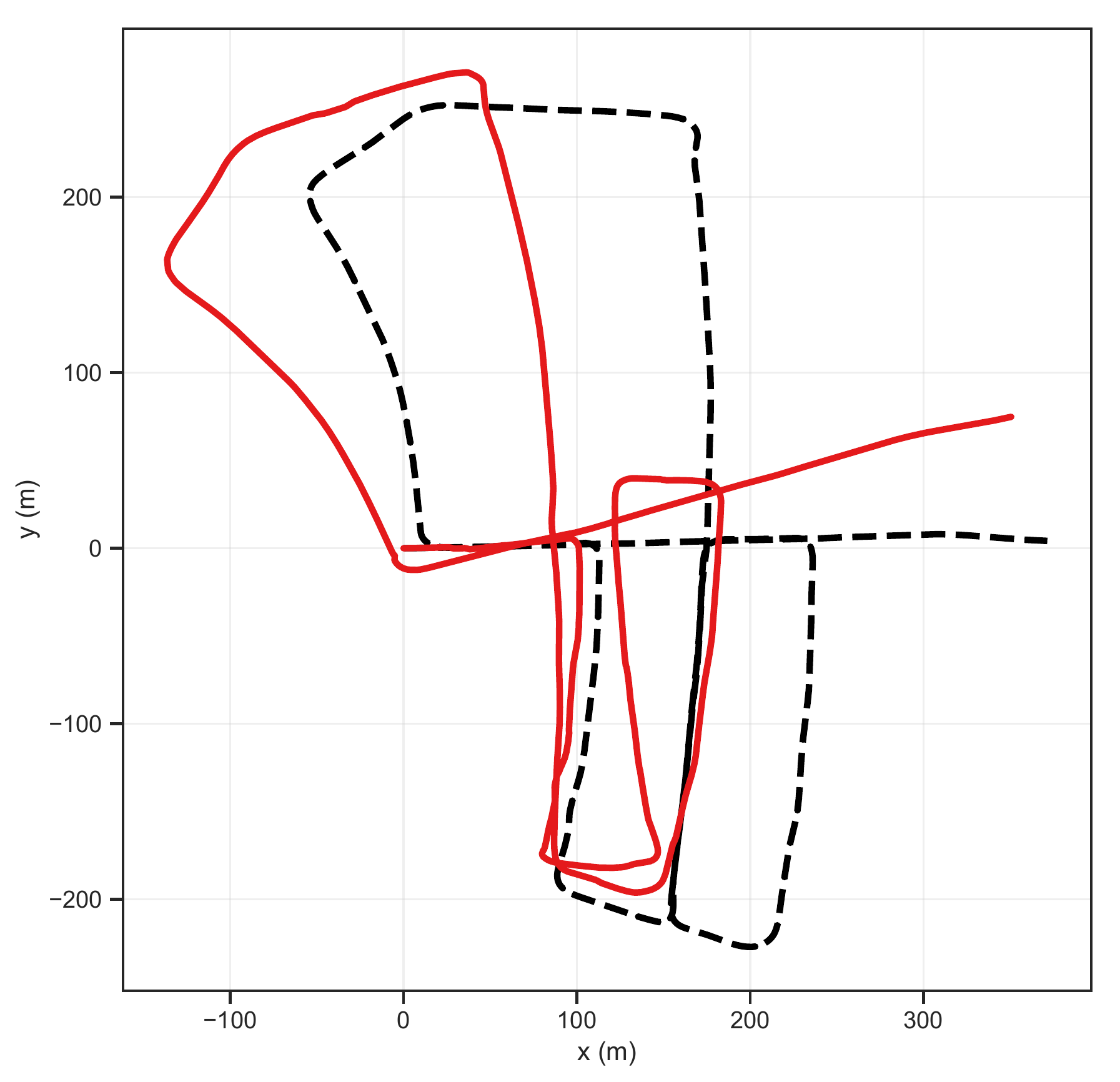}
    \caption{Max-Likelihood \label{fig:traj_error-ml}}
  \end{subfigure}%
  \begin{subfigure}{0.24\linewidth}
    \centering
    \includegraphics[width=1.0\linewidth]{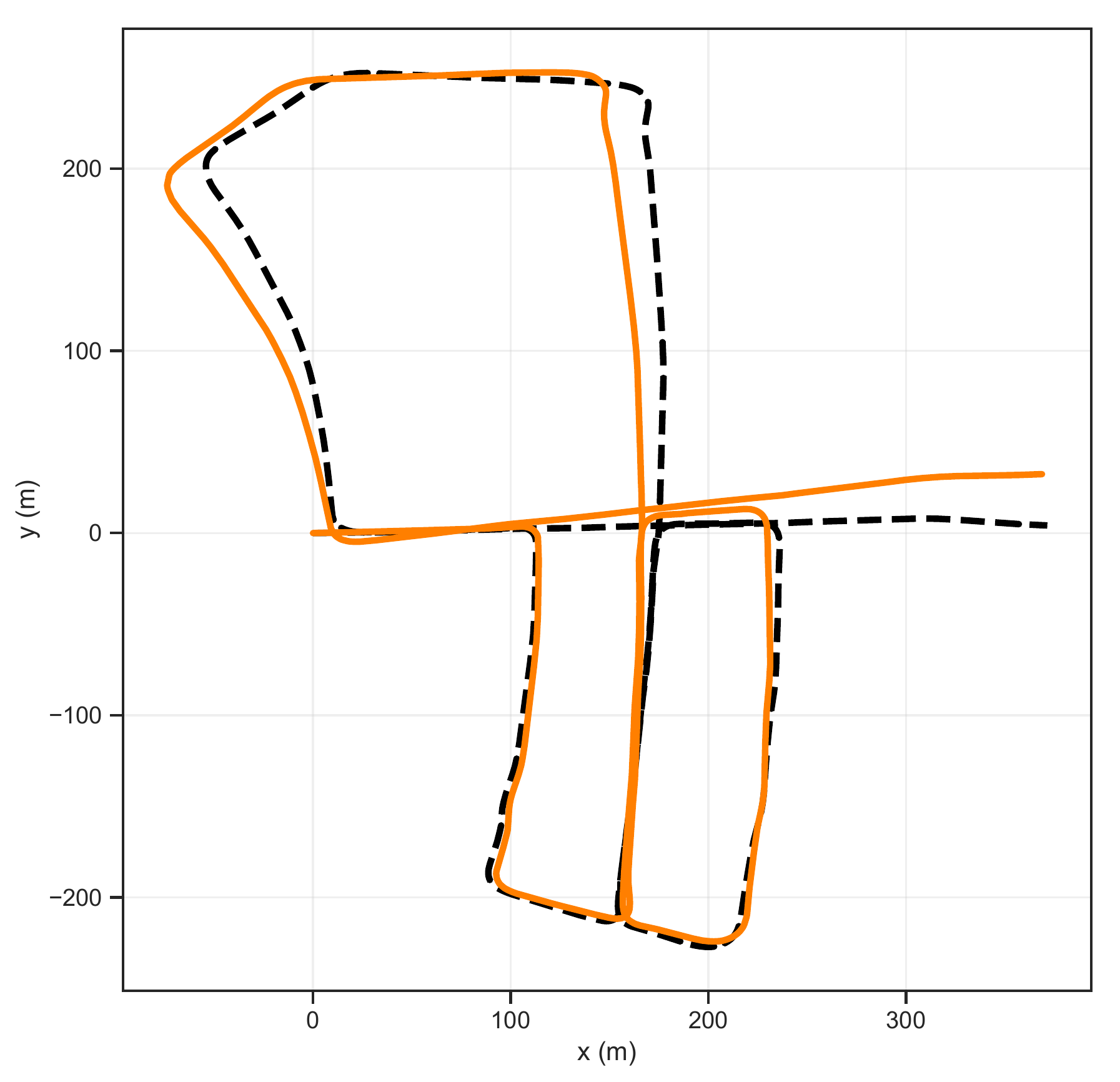}
    \caption{Gaussian PDA \label{fig:traj_error-gpda}}
  \end{subfigure}%
  \begin{subfigure}{0.24\linewidth}
    \centering
    \includegraphics[width=1.0\linewidth]{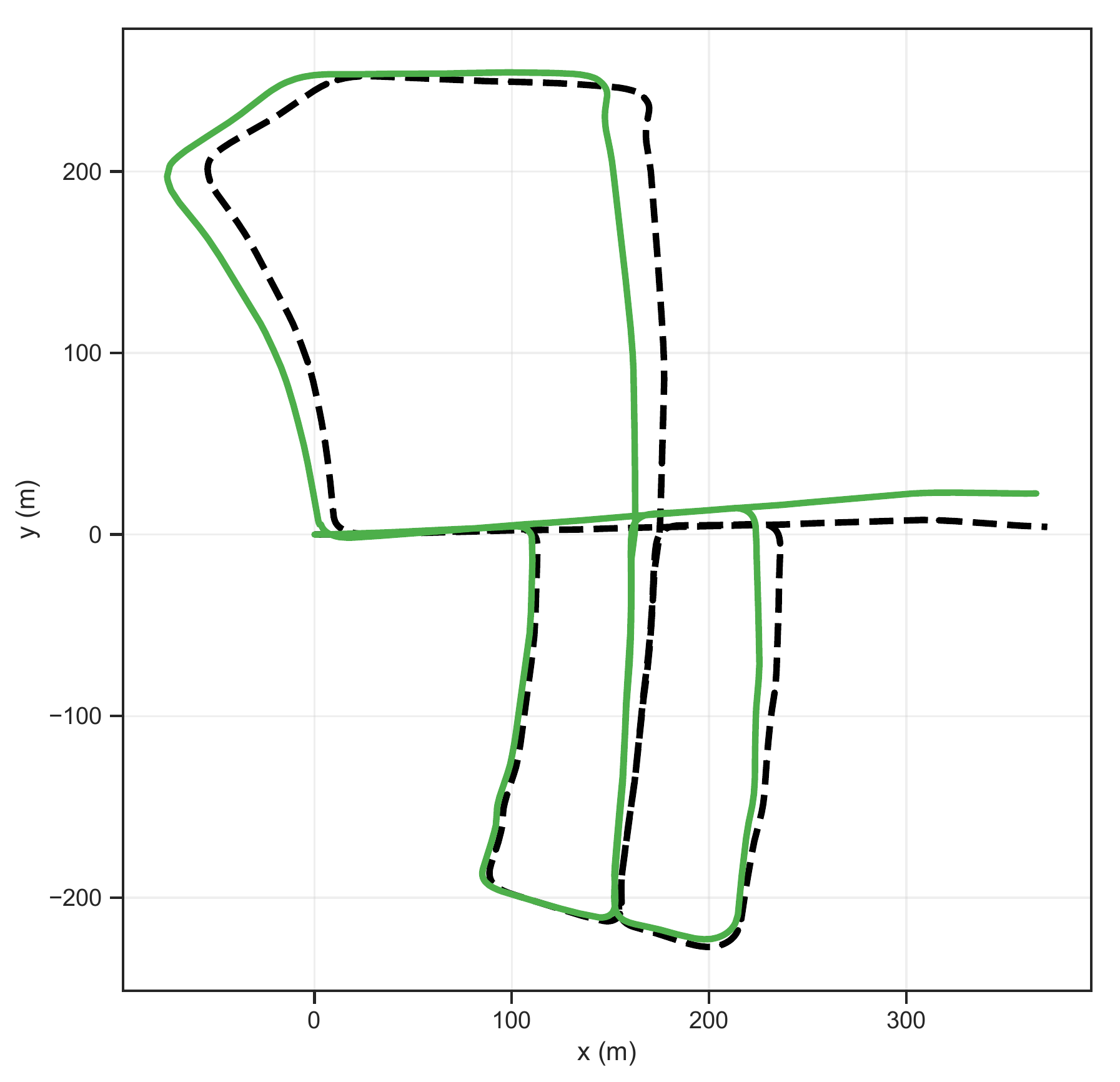}
    \caption{Semantic MM (Ours) \label{fig:traj_error-maxmixtures}}
  \end{subfigure}%
  \begin{subfigure}{0.24\linewidth}
    \centering
    \includegraphics[width=1.0\linewidth]{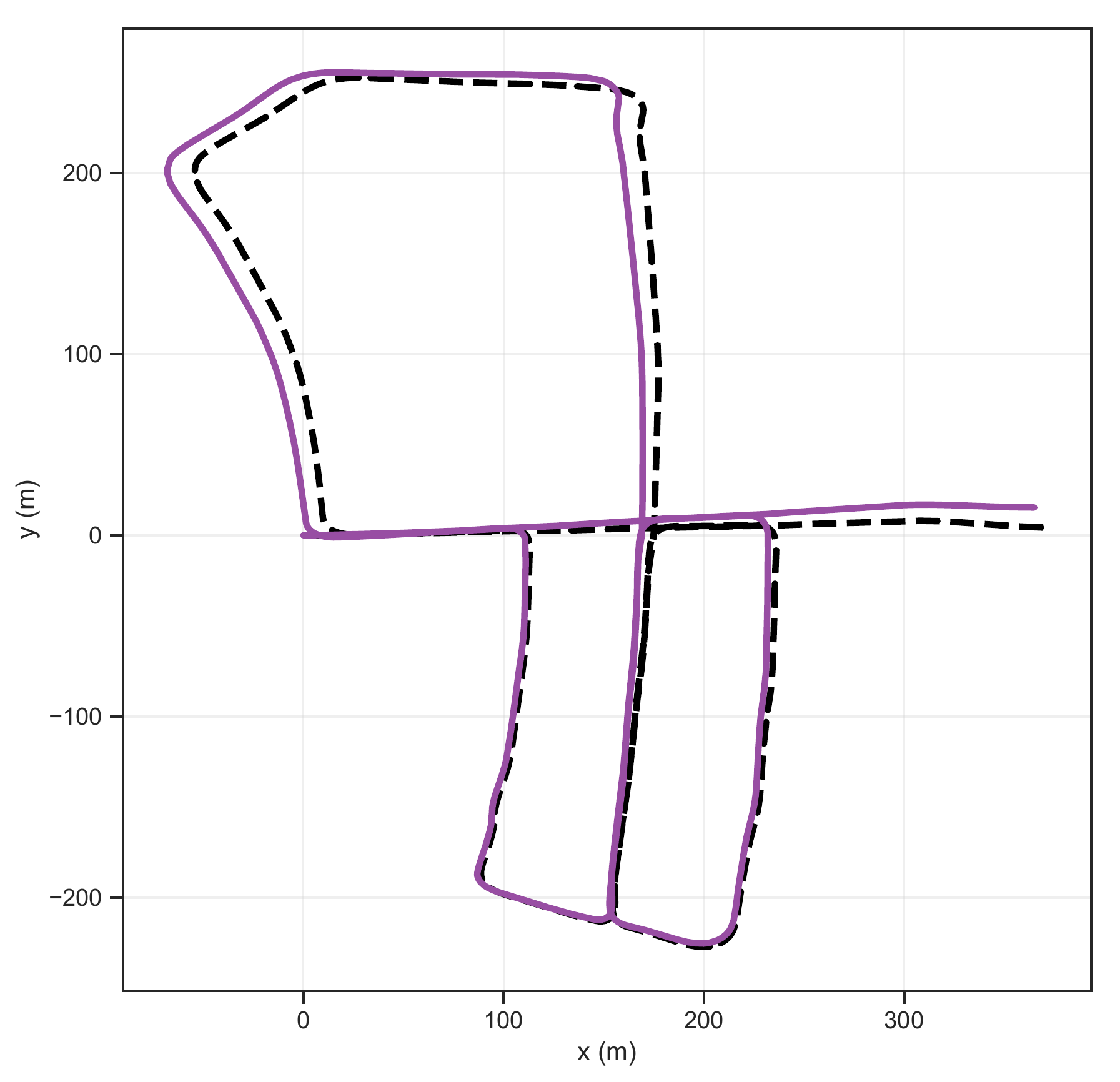}
    \caption{Semantic MM + NH (Ours) \label{fig:traj_error-maxmixtures_nh}}
  \end{subfigure}
  \caption{Estimated trajectories for KITTI Sequence 5. False loop closures
    cause the maximum-likelihood data association method to fail
    catastrophically, while all of the probabilistic methods show improved
    robustness. \label{fig:kitti-traj-error}}
\end{figure*}


All computational experiments were implemented in C++ using the Robot Operating
System (ROS) \cite{quigley2009ros} and the implementation of \textrm{iSAM2}
\cite{kaess2012isam2} within the GTSAM \cite{dellaert2012factor} library for
optimization and covariance recovery. We demonstrate our approach on 3D
visual SLAM tasks using data collected during indoor navigation with an MIT
RACECAR vehicle\footnote{\url{https://mit-racecar.github.io/}} equipped only
with a ZED stereo camera \cite{zed}, as well as with stereo image data from the
KITTI dataset \cite{geiger2012kitti, geiger2013vision}. Experiments were run on
a single core of a 2.2 GHz Intel i7 CPU. We use \textrm{evo} \cite{grupp2017evo}
for trajectory evaluation\footnote{We provide trajectory comparisons for all of
  the methods tested without landmarks visualized, but more detailed
  visualizations and videos can be found on the project page:
  \url{https://github.com/MarineRoboticsGroup/mixtures_semantic_slam}}.

In both experiments, we compared two variants of the proposed method: semantic
max-mixtures (MM) and semantic max-mixtures with null-hypothesis data
association (MM+NH) to a known data association (Known DA) baseline, na\"ive
maximum-likelihood (ML) data association (which makes a single association with
the landmark maximizing Eq. \eqref{eq:factored-da-likelihood}), as
well as an expectation-maximization approach similar to that of
\cite{bowman2017probabilistic}, here referred to as Gaussian probabilistic data
association (GPDA). We use a threshold on the marginal likelihood in
\eqref{eq:factored-da-likelihood} to determine new landmarks, as well as to
produce the set of landmark candidates. This is similar to standard Mahalanobis
distance-based thresholds, but considers also the semantic
likelihood\footnote{We use a $\chi^2$ test with confidence 0.9 and
  null-hypothesis weight of 0.1.}.

\subsection{MIT RACECAR Dataset}

We collected roughly 25 minutes of data during indoor navigation with the MIT
RACECAR mobile robot platform over a roughly 1.08 km trajectory. We sampled
AprilTag \cite{wang2016apriltag, mayluta2019long} detection keyframes at a rate of 1 Hz resulting in 702
observations of 262 unique tags. Odometry was obtained using the ZED stereo
camera visual odometry \cite{zed}. The use of AprilTags uniquely allows us to
obtain a baseline ``ground-truth'' solution with known data associations. We
artificially assigned semantic labels to each AprilTag by considering the true
tag ID modulo $C$ for a $C$-class semantic SLAM problem. In the experiments
presented in this paper, we use $C=2$ classes, as we found it to be one of the
most challenging situations\footnote{In general, with a ``good'' detector, the
  presence of many unique semantic classes among landmarks makes the data
  association problem easier.}. While AprilTags give generally accurate
orientation information, we typically cannot expect this of neural network-based
object detectors. For this reason, we set a large standard error on roll, pitch,
and yaw of AprilTag detections to prevent orientation information from giving
any substantial data association cues. Furthermore, this experimental setup
allows us to apply classification error and simulate additional odometry noise
in a repeatable way to study the trade off in data association performance with
noise in classification and odometry.

In Figure \ref{fig:apriltag-error}, we provide box-plots summarizing statistical
results of trajectory error on the MIT RACECAR dataset. Specifically, we
considered robustness to simulated additional odometry noise and detector
misclassification. We calibrated an initial odometry model, but in testing we
add simulated Gaussian noise of $x$, $y$, and yaw\footnote{We take the baseline
  simulated noise model as $\sigma_x = 1.5 \mathrm{mm}$, $\sigma_y = 0.75
  \mathrm{mm}$, and for yaw, $\sigma_{\gamma} = 0.00225\ \mathrm{rad}$ in robot
  frame.}, multiplied by a scale factor varying
from 0 to 10. We simulate classification error for a misclassification rate
$\alpha$ (from 0\% to 50\%) by sampling from a semantic measurement model with
confusion matrix equal to $1 - \alpha$ on the diagonal and $\alpha$ on the
off-diagonal. We find that while errors in all methods increase with added noise
in odometry and misclassification, all of the probabilistic methods generally
outperform maximum-likelihood. Most significantly, we find that the addition of
the null-hypothesis to our approach drastically reduces error in all tests.
Beyond the ability of the null-hypothesis method to reject bad loop closures,
the addition of the null hypothesis may help prevent the max-mixtures approach
from becoming ``stuck'' in a local optimum by decreasing the high cost
associated with being ``between'' hypotheses; a region that must be crossed
before hypothesis switching can take place. We contextualize these quantitative
results with qualitative trajectories plotted in Figure
\ref{fig:apriltag-traj_error} for a 2-class problem with 10\% misclassification
and 10\% additional odometry error.

\begin{table}[t]
  \centering
  \begin{tabular}{c c c c c}
    \hline
    Method & Max Error & Mean Error & Median Error & RMSE  \\\hline
    ML & 126.46 & 52.78 & 59.66 & 62.35 \\ 
    GPDA & 30.76  & 10.52  & 8.90 & 12.11 \\
    MM & 23.23 & 9.31 & 8.31 & 11.37 \\
    MM + NH & \bf{19.88} & \bf{6.48} & \bf{6.44} & \bf{7.70} \\ \hline
  \end{tabular}
  \caption{Translation error (m) on KITTI Sequence
    5. \label{table:kitti-trans-error}}
\end{table}

\begin{table}[t]
  \centering
  \begin{tabular}{c c c c c}
    \hline
    Method & Max Error & Mean Error & Median Error & RMSE  \\\hline
    ML & \bf{0.42} & 0.15 & 0.11 & 0.19 \\ 
    GPDA & 0.51 & 0.06 & 0.052 & 0.069 \\
    MM & 0.54 & 0.055 & 0.049 & 0.065 \\
    MM + NH & 0.58 & \bf{0.043} & \bf{0.037} & \bf{0.053} \\ \hline
  \end{tabular}
  \caption{Rotation error (rad) on the KITTI Sequence
    5. \label{table:kitti-rot-error}}
\end{table}

\subsection{KITTI Dataset}

We also evaluate our approach on stereo camera data from the KITTI dataset
odometry sequence 5~\cite{geiger2012kitti}. In our experiments, we use the
MobileNet-SSD object detector~(\cite{huang2017speed, howard2017mobilenets,
  liu2016ssd}), from which detections were obtained at approximately 10 Hz. We
threshold the confidence of the detector at 0.8, using detections of cars as
landmarks. We use VISO2 stereo odometry for visual
odometry~\cite{geiger2011stereoscan}. We estimate the range and bearing to cars
as the average range and bearing to all points tracked by VISO2 that project
into the bounding box for a given car detection. Despite this very noisy
landmark signal, we show qualitatively in Figure \ref{fig:kitti-traj-error} that
all of the probabilistic data association methods successfully recover
reasonable trajectory estimates, while the maximum-likelihood approach fails
catastrophically due to incorrect loop closures. These results are corroborated
by the translation and rotation errors, summarized in Tables
\ref{table:kitti-trans-error} and \ref{table:kitti-rot-error}, respectively.

\section{Conclusion and Future Work}

We have proposed an approach to semantic SLAM with probabilistic data
association based on approximate max-marginalization of data associations. This
led to a ``max-mixture''-type approach to factor graph SLAM, amenable to
nonlinear least-squares optimization. We have shown how mixture component
weights can be computed from joint semantic and geometric measurements, and
null-hypothesis associations can be incorporated to reject bad loop
closures. We evaluated the proposed approach on real stereo image data with
known data associations using AprilTags \cite{wang2016apriltag} under a variety
of simulated odometry and detection noise models, as well as on stereo image
data from the KITTI dataset using noisy detections of cars as landmarks. We have
shown that our approach is competitive with recent expectation-maximization
methods for data association in semantic SLAM and drastically outperforms the
common maximum-likelihood approach, particularly with use of null-hypothesis
data association, while being similarly easy to implement within existing factor
graph optimization frameworks. The particular benefits of the null-hypothesis in
the proposed framework suggest that the ability to reject incorrect loop
closures is a necessity for semantic SLAM systems relying on object detectors,
and this work is one step toward unifying existing literature in robust SLAM
with recent work on data association for object-level/semantic SLAM.

In this work we used classifications from object detectors to disambiguate data
associations, but alternative semantic descriptors can be modeled and
incorporated similarly, for example using neural network-based feature matching
techniques \cite{tian2017l2, yi2016lift, han2015matchnet}. Additionally, our
experience simulating semantic SLAM with AprilTags suggested that orientation
can provide very useful cues for data association. While we focused on
approaches that give limited geometric information about objects, methods like
\cite{nicholson2018quadricslam} that infer the full pose of objects may greatly
improve accuracy and robustness of data association.

        \addtolength{\textheight}{-4.5cm}   



    %

\bibliographystyle{IEEEtran}
\bibliography{main}
%
%
%
%
%
%
%

\end{document}